\ifdefined\ARXIV
\documentclass[letterpaper, 10pt, twocolumn]{article}
\usepackage[margin=0.75in]{geometry}
\usepackage{libertine}
\renewcommand\footnotemark{} 
\date{}
\else
\documentclass[letterpaper, 10 pt, conference]{ieeeconf}
\IEEEoverridecommandlockouts
\fi

\usepackage[margin=0.75in]{geometry}
\usepackage{hyperref}
\hypersetup{colorlinks = true, citecolor = {black}, linkcolor = black}
\usepackage[USenglish]{babel}

\usepackage{amsmath}
\usepackage{amssymb}
\usepackage{babel}
\usepackage{graphicx}
\usepackage{subfig}
\usepackage{url}
\usepackage{algorithm}
\usepackage{algpseudocode}
\algnewcommand{\Initialize}[1]{%
  \State \textbf{Initialize:}
  \Statex \hspace*{\algorithmicindent}\parbox[t]{.8\linewidth}{\raggedright #1}
}

\usepackage{cite}

\usepackage{xcolor}   
\newcommand{\rev}[1]{{\color{black} #1}} 
\newcommand{\revv}[1]{{\color{black} #1}} 

\title{\LARGE \bf Towards High-Payload Admittance Control for Manual Guidance with Environmental Contact}

\author{Kevin Haninger$^*$, Marcel Radke$^*$, Axel Vick, J{\"o}rg Kr{\"u}ger
\thanks{* Equal contribution, alphabetical order. Authors affiliated with the Department of Automation at the Fraunhofer Institute for Production Systems and Design Technology. Email: \texttt{firstname.lastname@ipk.fraunhofer.de}. 
}
\thanks{This project has received funding from the European Union's Horizon 2020 research and innovation programme under grant agreement No  820689 — SHERLOCK.}
}

\begin{document}

\maketitle
\begin{abstract}
  Force control enables hands-on teaching and physical collaboration, with the potential to improve ergonomics and flexibility of automation. Established methods for the design of compliance, impedance control, and \rev{collision response} can achieve free-space stability and acceptable peak contact force on lightweight, lower payload robots. Scaling collaboration to higher payloads can allow new applications, but introduces challenges due to the more significant payload dynamics and the use of higher-payload industrial robots.
  
  To achieve high-payload manual guidance with contact, this paper proposes and validates new mechatronic design methods: standard admittance control is extended with damping feedback, compliant structures are integrated to the environment, and a contact response method which allows continuous admittance control is proposed. These methods are compared with respect to free-space stability, contact stability, and peak contact force.  The resulting methods are then applied to realize two contact-rich tasks on a 16 kg payload (peg in hole and slot assembly) and free-space co-manipulation of a 50 kg payload. 
\end{abstract}

\section{Introduction}
\begin{figure}
    \centering
    \includegraphics[width = 0.9\columnwidth]{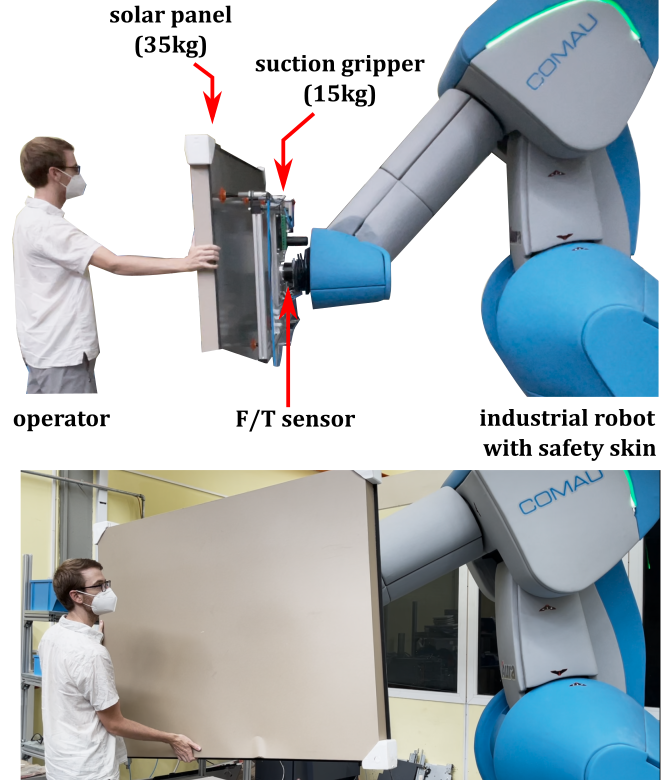} 
    \caption{Manual guidance with a heavy payload (50 kg) on a COMAU AURA robot, with 170 kg payload capacity and safety touch sensitive skin.}
    \label{fig:setup_man_guide}
\end{figure}

Physical interaction with a robot allows intuitive position teaching and collaboration, which can improve task programming and online flexibility.  Force control, impedance control, and other variants \cite{whitney1987, Hogan.1985_partItheory, keemink2018} can enable safe physical human-robot interaction (HRI), allowing robot position to be modified by human or environmental forces, which can reduce cycle time \cite{lecours2012} and human exertion \cite{dimeas2016}. 

However, contact couples the robot with unknown environment or human dynamics, where stability and maximum force bounds should be maintained. Robots with joint torque sensing can implement impedance control with inner-loop torque control  \cite{albu-schaffer2007}, reducing rendered impedance to improve safety and transient contact force. However, the payload currently achievable with joint torque sensing is limited -- commercially, to 14 kg. Impedance control via current monitoring, where joint torque is estimated from motor current, also does not scale to the higher gear reductions currently required for high payloads, as gearbox friction dominates external forces.  

While robots for physical HRI today have limited payload and reach, there are benefits to higher-payload collaboration. The ergonomic benefits of HRI increase at higher payloads, where the robot can reduce the weight carried by the human \cite{gosselin2013}.  Robot support of bulky objects can allow tasks which previously required two workers to be done with one \cite{franceschi2020, roveda2018c}. However, increased payload inertia introduces challenges for responsiveness, stability, and safety \cite{Aghili.2010, lecours2012, audet2021, surdilovic2020}.  

Human safety is critical, especially at higher payloads and with high impedance industrial robots. Hands-on-payload interaction makes it difficult to use an enabling switch (which the human must press to enable motion), so safety via power and force limiting must play a larger role.  Here, methods for the reduction of transient and quasi-static contact forces are proposed and compared, towards verifiable power and force limiting.  However, additional aspects for certification (e.g. Performance Level of the sensor and communication protocols) are outside the scope of this paper. 

High payload HRI has been demonstrated by separating human input from the payload with either a separate force/torque (F/T) sensor \cite{lecours2012, surdilovic2020} or a low-impedance input channel \cite{audet2021}. However, direct co-manipulation of the payload as seen in Figure \ref{fig:setup_man_guide} simplifies the system, allowing a single sensor for environmental contact, human input, and collision detection, as well as increasing the flexibility in how the operator grips the payload \cite{roveda2018c}.  The ability to set the rotation center or virtual constraints can also improve ergonomics and task performance \cite{bowyer2014a}.

Using an industrial robot for admittance control can simplify mechanical integration, but the high-impedance hardware and inner-loop position control makes contact stability more difficult \cite{colgate1989} and increases collision forces \cite{haddadin2017, haninger2019a}. Collision is especially critical with larger payloads as it provides impulsive input with a broad frequency range which can excite payload resonance modes. \rev{ Many control architectures proposed for high-payload co-manipulation do not experimentally test contact \cite{sidiropoulos2021, keemink2018}.}

These issues can be partially addressed through control. Peak contact force and contact stability can be improved through feedforward control, which can be the inverse of the robot dynamics \cite{surdilovic2007a}, a PID compensator \cite{cao2020} or a lead controller \cite{keemink2018}. In hands-on-payload applications, the compensation of inertial payload forces has been proposed to increase sensitivity \cite{Aghili.2010, keemink2018, S.Farsoni.2017}. Alternative force control architectures can be used, such as disturbance observer-based control \cite{katsura2007} or an arbitrary finite impulse response force controller \cite{pham2020}. However, standard admittance control has well-studied stability properties and allows methods to adapt the admittance parameters online \cite{abu-dakka2020, ficuciello2015}.  

However, control has limitations.  The ability to reduce payload oscillation through control is limited  as well as the ability to reduce the transient contact force \cite{haddadin2017}. As such, this paper also considers the use of collision detection and environmental compliance. While environmental compliance is often used informally in admittance control \cite{katsura2007, ferraguti2019}, we show it can be integrated into compliant tables to allow practical contact-rich tasks. 

This paper identifies and partially addresses new challenges of high-payload co-manipulation, mostly resulting from the significant payload inertia and resulting lower frequency payload resonance. Compared to published hands-on-payload co-manipulation \cite{roveda2018c, Aghili.2010} (14 and 12 kg payloads), we demonstrate co-manipulation up to a 50 kg payload. Compared to state of the art admittance control \cite{keemink2018}, we add damping feedback, collision response, and validate with physical experiments. First, the dynamic model for robot, payload and control architecture are proposed and analyzed. A collision response method which allows continuous co-manipulation in contact is proposed. Contact experiments compare \rev{the proposed methods with established variable admittance control and payload acceleration compensation}. Two contact-rich co-manipulation tasks are demonstrated (double peg-in-hole and a slot assembly task) with a 16 kg payload, and free-space manipulation of a 50 kg payload.  \rev{Finally, design conclusions are drawn for high-payload admittance control with contact.}

\section{Admittance Architecture, Analysis, and Implementation}
This section introduces the dynamic model, admittance controller, and analyzes passivity and contact resonance.  

\subsection{Control Architecture}
\begin{figure}
    \centering
    \includegraphics[width = \columnwidth]{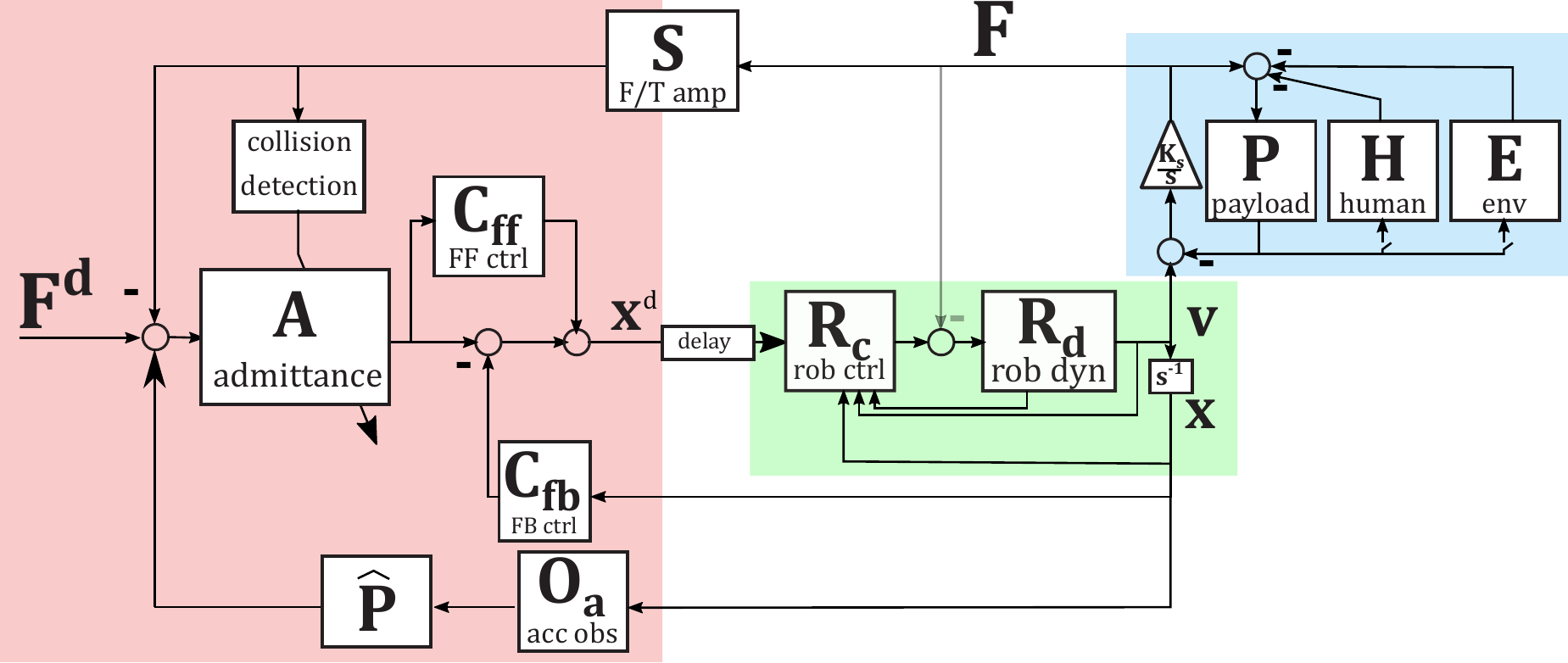}
    \caption{Admittance control architecture, where the implemented controller (red) sends commands to a position controlled robot (green). The robot has a F/T sensor and gripper with stiffness $K_s$ which couples the payload $P$ to the robot.}
    \label{fig:admittance_control_architecture}
\end{figure}

The controller, robot dynamics and environment can be seen in Figure \ref{fig:admittance_control_architecture}, shown for a single DOF in operational space. The rotational DOFs realize the desired admittance about the X/Y/Z axis of the TCP, and $F$ is transformed into this coordinate system before payload compensation, both of which occur in the block $S$.

In green is the closed position-controlled robot, which receives TCP position commands $x^d$ at $1250$ Hz and uses a classical cascade position controller with hierarchical position/velocity/current control loops. We simplify the robot controller $R_c$ and dynamics $R_d$ by considering the closed-loop robot dynamics \rev{in Laplace domain} as $R(s) = V(s)/X^d(s)$, where the physical impact of forces $F$ on the robot are ignored, assumed to be rejected by the high-gain position control.

The environment can be seen in Figure \ref{fig:admittance_control_architecture} in blue, where a stiffness $K_s$ couples a payload $P=(M_ps+B_p)^{-1}$ to the robot flange. The environment $E$ and human $H$ apply forces to the payload in parallel \rev{and are both shown in Figure \ref{fig:admittance_control_architecture} for completeness.  In the sequel we consider $E$ as a pure stiffness ($E=K_e/s$), and do not consider $H$ in closed-loop analysis as typical human arm stiffnesses are $<3$ N/mm \cite{tsumugiwa2002}, so are typically dominated by environment stiffness (here, $>20$ N/mm).}

We extend the control architecture of \cite{keemink2018} with a feedback controller from robot position $C_{fb}$ as an additional damping term \rev{based on the real robot velocity}. We note that damping in the admittance controller is based on the virtual velocity of the admittance integrator ($\dot{x}^d$), which, due to delay and robot dynamics, does not correspond to the real robot velocity. However, damping \rev{based on the real robot velocity} is critical for passivity and stable interactive control \cite{colgate1997}. The following control blocks \rev{from Figure \ref{fig:admittance_control_architecture}} are defined \rev{in Laplace domain} as:
\begin{align}
    A(s) & = \frac{1}{M_as^2+B_as} \label{eq:admittance}\\
    C_{ff}(s) & = K_l s \\
    C_{fb}(s) & = B_{fb}\frac{\omega_v s}{s+\omega_v} \label{eq:cfb}\\
    O_a(s) & = \frac{\omega_v \omega_as^2}{(s+\omega_a)(s+\omega_v)} \label{eq:oa}
\end{align}
where \rev{parameters} $M_a$ and $B_a$ are the admittance mass and damping, $K_l$ a lead controller gain, $B_{fb}$ a feedback damping term, $\omega_v$ and $\omega_a$ filter constants for the velocity and acceleration estimation. \rev{The Figure \ref{fig:admittance_control_architecture} block} \rev{$\hat{P}=M_p$} is the estimated payload inertia. 

The following transfer functions can be written, omitting the argument $(s)$ for compactness:
\begin{align}
    \frac{V}{F} = G & = -\frac{A(1+C_{ff})R}{1-A(1+C_{ff})R\hat{P}^{-1}O_as^{-1}+RC_{fb}s^{-1}}, \label{ol_dynamics} \\
    \frac{F}{V} = \overline{E} & = \frac{K_ss^{-1}(1+PE+PH)}{1+PE+PH+PK_ss^{-1}}, \\
    \frac{F}{F^d}  & = \frac{-G\overline{E}}{1+G\overline{E}} \label{cl_dynamics}.
\end{align}

\subsection{Controller Design}
The dynamics of \eqref{ol_dynamics} and \eqref{cl_dynamics} will be used for controller design. Stability with environmental contact is a foundational problem for admittance control \cite{hogan1988}, and various stability conditions have been used: passivity  \cite{Hogan.1985_partItheory, keemink2018}, mixed small-gain/pasisivity \cite{haninger2018b}, or assuming environmental dynamics and showing closed-loop stability \cite{eppinger1992, buerger2006}. 

\begin{figure}
    \centering
    \includegraphics[width=0.7\columnwidth]{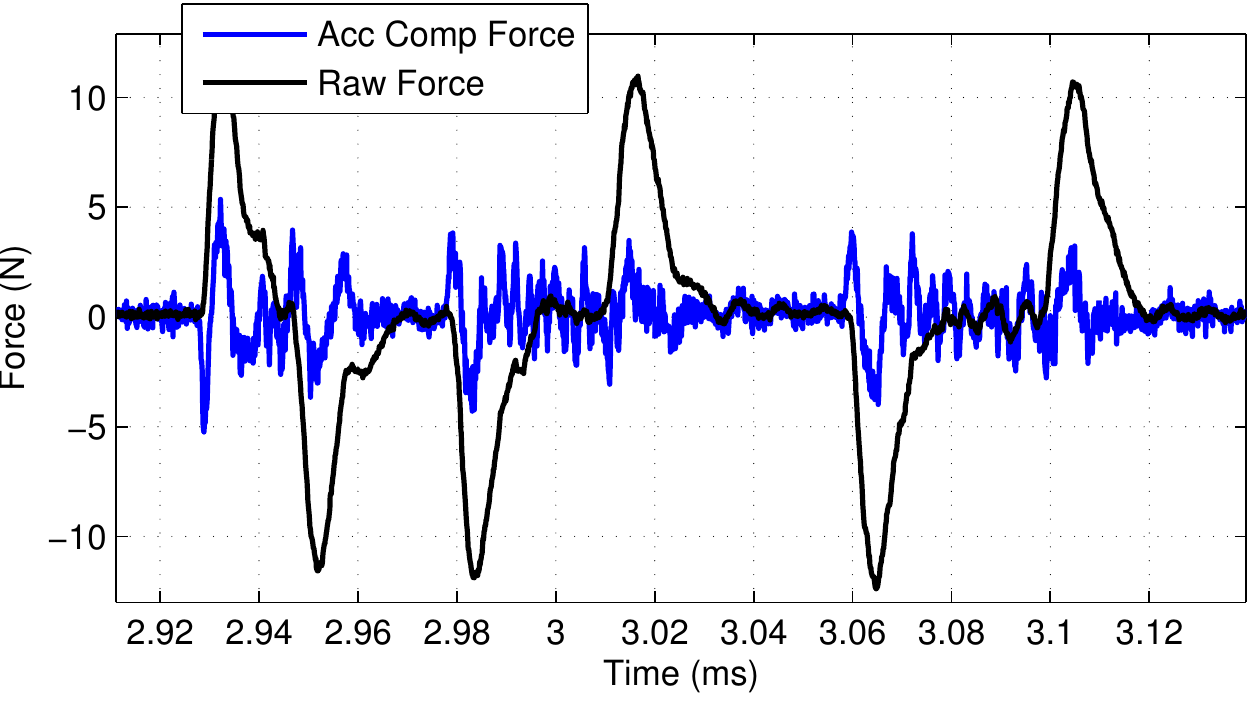}
    \caption{Force measurements during jogging, with and without compensation of payload acceleration. The payload compensation substantially increases noise in the force, from RMS value of $0.14$ to $0.40$ N at rest.}
    \label{fig:acc_comp}
\end{figure}

In the following analysis, control parameters are $M_a = 10$, $B_a = 1000$, $K_l = 0.02$, $B_{fb} = 2E^ {-5}$, $\omega_a = 200$, $\omega_v=120$, and dynamic parameters $K_s = 3E^5$, $E=5E^5/s$, $H=0$, $P=(16s + 630)$ and $R=(100s^2+400s)(5s^2+120s+400)^{-1}$, where the robot dynamics are the derivative of a well-damped position-controlled inertia with $3$ Hz bandwidth. An input time delay of $T_d = 2.4e-3$ is considered on $R$. These values of $\omega_a$ and $\omega_v$ reflect those which allowed acceptable noise characteristics on the hardware for the acceleration compensation, seen in Figure \ref{fig:acc_comp}. Idealized payload acceleration compensation is considered where $O_a = s^2$ instead of with a low-pass filter as in \eqref{eq:oa}.  

\subsubsection{Rendered admittance in free space}
\begin{figure}
    \centering
    \subfloat[Measured force $F$ to robot velocity $v$ in free space ($E=H=0$) with delay $T_d = 2.4E^{-3}$. \label{bode_free_space}]{\includegraphics[width=0.9\columnwidth]{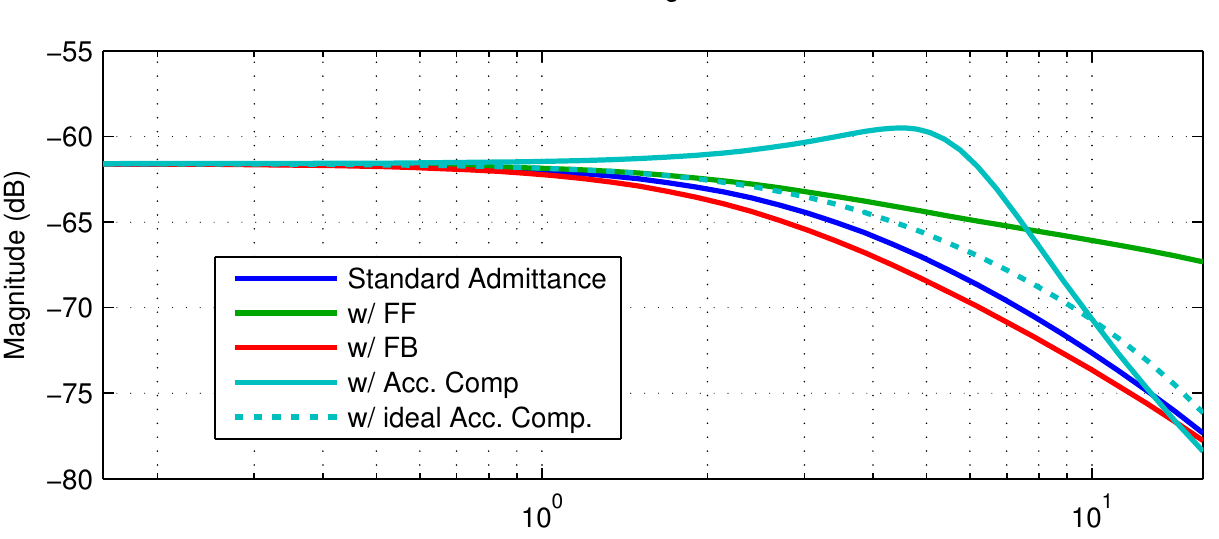}}\\
    \subfloat[Desired force $F^d$ to force $F$ in contact ($E=5E^{+5}/s, H=0$) \label{bode_contact}]{\includegraphics[width=0.9\columnwidth]{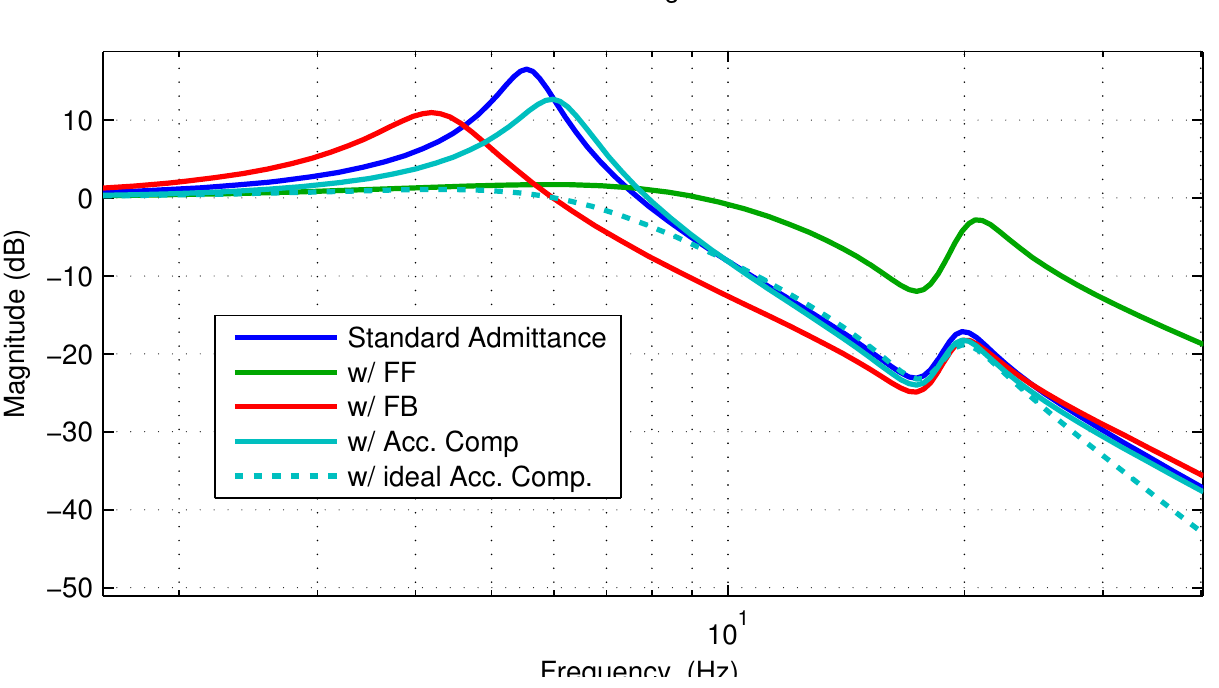}} 
    \caption{Bode magnitude plot for transfer function $F\rightarrow v$ in free space (a) and $F^d\rightarrow F$ in environmental contact (b) under various controllers. }
    \label{fig:bode_plots}
\end{figure}

The rendered admittance of the various controllers are compared with $E=H=0$, comparing $F\rightarrow V$ as seen in Figure \ref{bode_free_space}. From the magnitude plot of Figure \ref{bode_free_space}, it can be seen that the lead compensator increases the admittance and ideal payload acceleration compensation increases it further, improving responsiveness. \rev{The damping feedback controller decreases admittance in the range 1-20 Hz, reducing responsiveness.}


\subsubsection{Force tracking in contact}
We then compare the magnitude of $F^d\rightarrow F$ when $E=5E^5/s$ as seen in Figure \ref{bode_contact}. Standard admittance control has a high resonance peak, which is eliminated by the feedforward lead controller and slightly reduced by the feedback controller. Ideal payload acceleration compensation reduces resonance, but the low-pass filtered version \rev{is not as effective}. From this, we conclude that the benefits of the acceleration compensation depend on the bandwidth of the acceleration estimation.

\rev{Contact stability is checked at a range of parameter values, and the impact of the controller components on minimum stable admittance damping $B_a$ can be seen in Fig. \ref{fig:resonance}(a) over a range of environment stiffness.  At higher $K_{e}$, payload compensation requires higher damping, but the feedback and feedforward control allow lower stable $B_a$, improving contact performance.}
\begin{figure}
    \centering
    \subfloat[$16$ kg payload, instability with admittance control and $K_l = 0.025$]{\includegraphics[width = 0.6 \columnwidth]{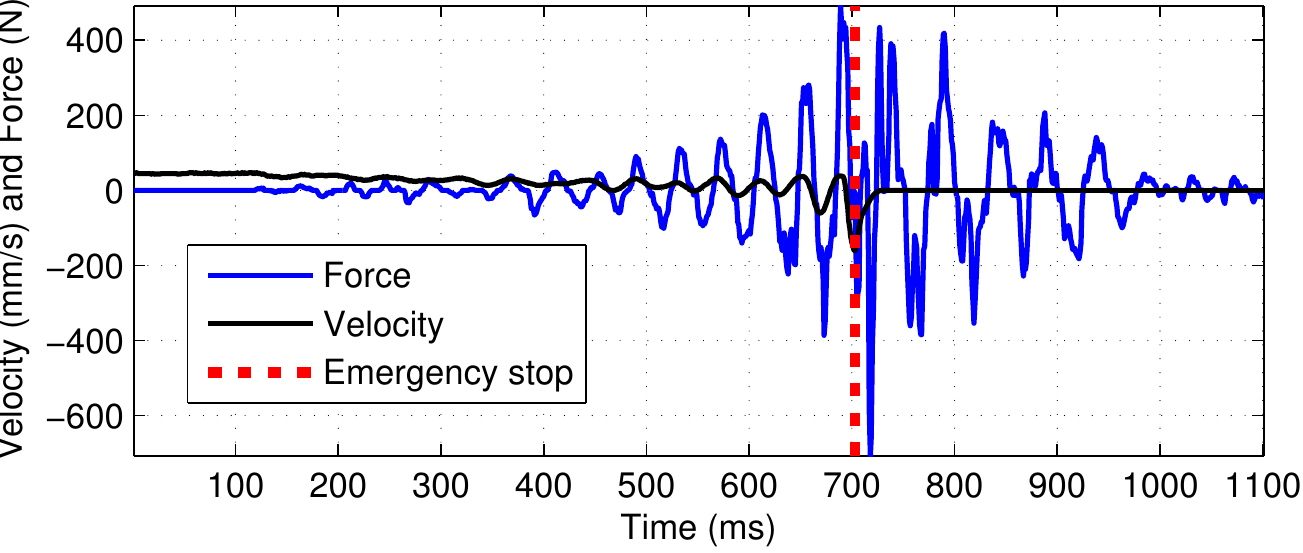}}\\
    \subfloat[$50$ kg payload, force oscillation during jogging]{\includegraphics[width = 0.6 \columnwidth]{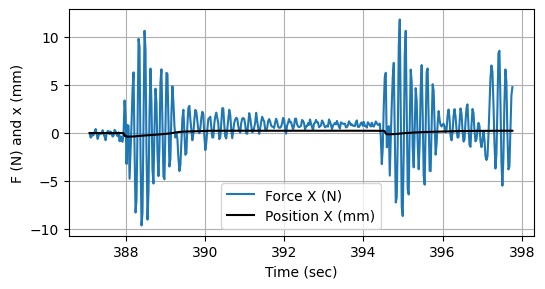}}
    \caption{Payload oscillation (a) admittance control with $K_l=0.025$ and $16$ kg payload, unstable mode at $25$ Hz (b) during jogging with $50$ kg payload, dominant mode at around $9$ Hz.}
    \label{fig:instability_kl}
\end{figure}

\begin{figure}
    \centering
    \subfloat[]{\includegraphics[width = 0.8\columnwidth]{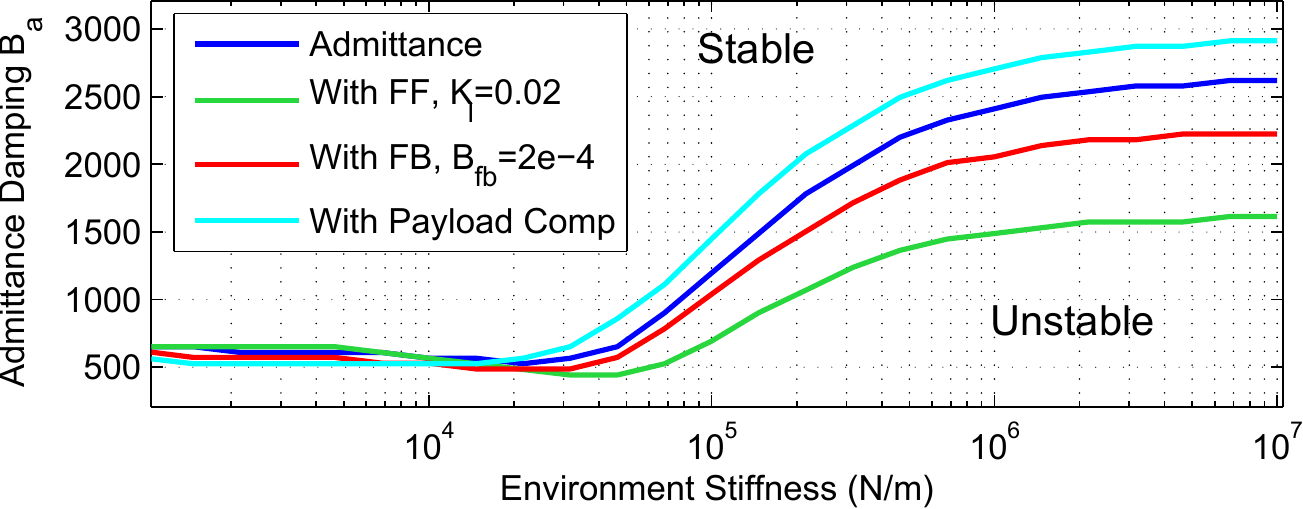}}\\
    \subfloat[]{\includegraphics[width = 0.8\columnwidth]{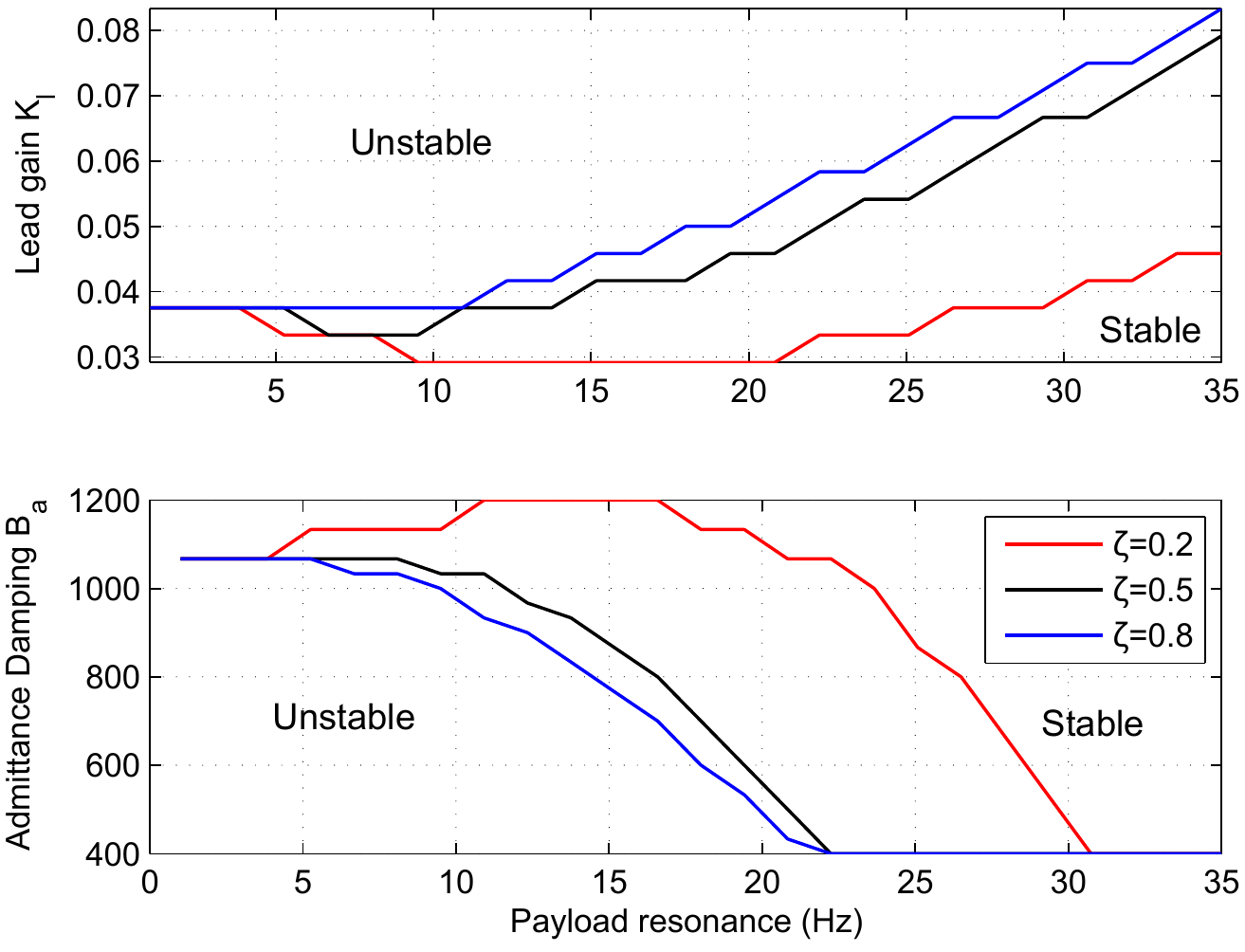}} 
    \caption{Impact of \rev{(a) Environment stiffness and controller components on minimum stable damping $B_a$ and (b)} payload natural frequency $\omega = \sqrt{M_p/K_s}$ and corresponding damping ratio $\zeta$ on stable parameter bounds.}
    \label{fig:resonance}
\end{figure}

\subsubsection{Instability from high lead gain}
While lead gain $K_l$ improves both free-space admittance and contact resonance, it can drive instability due to the resonance in the payload, even in free space as seen in Figure \ref{fig:instability_kl}(a). This instability does not appear analytically from passivity analysis of $F\rightarrow V$, but it can be found through classical linear stability when delay and payload resonance are modelled. We note that this problem is similar to that of non-collocation \cite{hogan1988, eppinger1992}, but here the resonance is after the sensor, not between sensor and actuator. 

The impact of payload resonance \rev{on stability bounds} is shown in Figure \ref{fig:resonance}, where payload mass and damping $M_p, B_p$, $P=M_ps+B_p$ are varied to achieve the resonant natural frequency $\omega=\sqrt{M_p/K_s}$ and damping ratio $\zeta=B_p/(2\sqrt{M_pK_s})$. The limit of stable lead gain $K_l$ and damping $B_a$ are found numerically, checking linear stability of \eqref{cl_dynamics} in contact with $E=5E^5/s$. As resonant frequency decreases, the minimum stable damping increases and the maximum stable lead gain decreases, both limiting performance.  This trend is exaggerated if the payload is poorly damped (red). Thus, the low-frequency and lightly-damped payload dynamics typical with a higher payload set limits on the stable admittance parameters. 

\subsection{Controller Implementation}
The transfer functions \eqref{eq:admittance}-\eqref{eq:oa} are discretized with a Tustin transformation at a sample time of $h = 8E^{-4}$, giving the following equation for admittance and feedforward controller
\begin{equation}
    \begin{split}
 (1+\frac{B_ah}{2M_a})x^d_{t+1} = 2x^d_{t}-(1-&\frac{B_ah}{2M_a}) x^d_{t-1}\\
 +\frac{h^2}{4M_a}\left(\left(1+2h^{-1}K_{l}\right)F_{t+1}\right.&\left.+2F_{t}+\left(1\texttt{-}2h^{-1}K_{l}\right)F_{t\texttt{-}1}\right),
  \label{discretized_adm}     
    \end{split}
\end{equation}
with discrete time samples $x^d_t$ and $F_t$. Note that Euler discretization was found to have worse contact stability, as shown in Figure \ref{fig:contact_tests_3}.  

An ME-Me\ss systeme KD6110 F/T sensor is used, with rated limits of $5$ kN and $250$ Nm. The GSV-8DS amplifier is used, with an internal analog low-pass filter with a cutoff frequency of 2500 Hz. A deadband of $5$ N and $1$ Nm is applied after the payload gravity compensation - all forces below this value are set to zero. This deadband keeps small gravitational compensation errors from causing robot motion. The RMS sensor noise at rest is $0.14$ N without acceleration compensation. 

\section{Collision Detection\label{sec:collision_detect}}
Contact can be made with pure force control, where the contact forces are regulated according to the admittance, reducing the quasi-static force in collision.  Alternatively, contact can be detected and used to change robot behavior, which has additional complexity but can reduce peak contact force \cite{haddadin2017, haningerc}.  

There are two challenges in co-manipulation contact detection. First, human and environmental contact forces must be distinguished \cite{haddadin2017}, where a common strategy is to assume human input is mostly low frequency, and use a threshold on a high-pass filtered force signal to distinguish contact \cite{ferraguti2019}. A second challenge is that in co-manipulation, often the task is not finished at the first contact - fine positioning must still be carried out. Thus, the collision response of \rev{a permanent robot stop} may be unsuitable.  In co-manipulation, the collision response of increasing of damping has been proposed \cite{okunev2012, dimeas2016, ferraguti2019}, which allows continuous co-manipulation.  However, this has a minimal impact on the initial peak contact force, which can be critical in high payloads.

\begin{algorithm}
\caption{Proposed collision response \label{coll_response}}
\begin{algorithmic}
    \Initialize{$\tt was\textunderscore contact = False$}
    \While {true}
    \State ${\tt contact} = F_{t+1}^j\mathtt{sign}(v^j) > \overline{F}  \,\,{\tt for \,\,any}\,\, j=1\dots3$
    \If{$\tt contact\, and\, not\, was\textunderscore contact$}
    \State $x^{d,k}_{t+1} \leftarrow x^{d,k}_t$  {\tt for} $k=1\dots 6$
    \State $x^{d,k}_{t-1} \leftarrow x^{d,k}_{t}$ {\tt for} $k=1\dots 6$
    \State $\tt was\textunderscore contact = True$
    \Else
    \State $x^{d,k}_{t+1} \leftarrow$ \eqref{discretized_adm} {\tt for} $k=1\dots 6$
    \EndIf
    \EndWhile
\end{algorithmic}
\end{algorithm}
To examine the continuity of this response, consider $x^d_t = x^d_{t-1}+\delta$, then \eqref{discretized_adm} simplifies to
\begin{equation}
    x^d_{t+1} = x^d_t + \frac{2M_a-B_ah}{2M_a+B_ah}\delta + g(f_{t+1},f_t,f_{t-1}),
\end{equation}
where $g(\cdot)$ is the contribution from forces and is unchanged from \eqref{discretized_adm}.  It can be seen that when $M_a>\frac{1}{2}B_ah$, a positive $\delta$ is carried forward to the next step due to the admittance integrator.  When $\delta=0$, this term drops and $x^d_{t+1}=x^d_t+g(f_{t+1},f_t,f_{t-1})$, so setting $\delta=0$ by modifying $x^d_{t-1}$ does not cause a discontinuity.     

\section{Contact Experiments\label{sec:controller_component_study}}
To validate the controller analysis, contact experiments are done without human contact to improve repeatability. Unless otherwise noted, experiments use linear mass and damping of $M_a = 13$, $B_a= 1000$, $K_l = 0.013$, and $B_{fb} = 7.5E^{-6}$, and contact is made by setting $F^d = 30$ N in the direction of desired contact.

\subsection{Controller Component Study } 
The impact of the feedforward/feedback controllers are compared in contact with a compliant table, seen with yellow foam in Figure \ref{fig:contact_tests_setup}.  

In Figure \ref{fig:contact_tests_1}, the improvement of the lead and feedback controllers can be seen - they reduce peak contact force and have reduced resonance, as predicted in Figure \ref{bode_contact}. The acceleration compensation is shown in Figure \ref{fig:contact_tests_3}, where the damping is decreased from acceleration compensation, and the peak contact force slightly increased, again as predicted by Figure \ref{bode_contact}.

\subsection{Collision Detection}
Contact responses with and without the collision response method proposed in Section \ref{sec:collision_detect} are carried out on a smaller payload robot (a COMAU Racer 7, with a 6.3 kg payload attached), reducing damping to $B_a=800$. The collision detection threshold of $\overline{F} = 7$ N is used, and the reference force of $F^d=30$ N is applied for the no collision detection and proposed collision detection. Additionally, the collision response of making a full stop, where $x^d_t$ is not updated after detection (thus ignoring $F$ and $F^d$ after collision).  \revv{The full stop keeps the robot motion control active, but keeps the position command $x^d$ at the same value. This halts the external} admittance control, which is typically unsuitable for co-manipulation, and is thus included for only comparison.

The results can be seen in Figure \ref{fig:collision_response_comparison}, where the proposed collision detection reduces peak contact force from $44$ to $36$ N. The admittance desired velocity $\dot{x}^d$ is also displayed, calculated by the numerical difference of $x^d_t$ and smoothed with a moving average filter of length $5$.  The immediate zeroing of $\dot{x}^d$ in contact can be seen, and the continuity in $x^d$ is maintained. As can be seen in Figure \ref{fig:collision_response_comparison}(a), the full stop eliminates force overshoot and oscillation, but cannot be continuously moved after contact without resetting the control. The improvement of the collision detection depends on the control parameters and environment stiffness.  When $K_l$ is increased from $0.013$ to $0.02$ in (b), there is almost no force overshoot to reduce.  When the environment is made more compliant, $K_e$, from $32$ to $15$ N/mm in (c), the collision detection improves performance, as the lower stiffness takes more time to build forces to stop the pure force control system.

\subsection{Environment Compliance}
\begin{figure}
    \centering
    \includegraphics[width = \columnwidth]{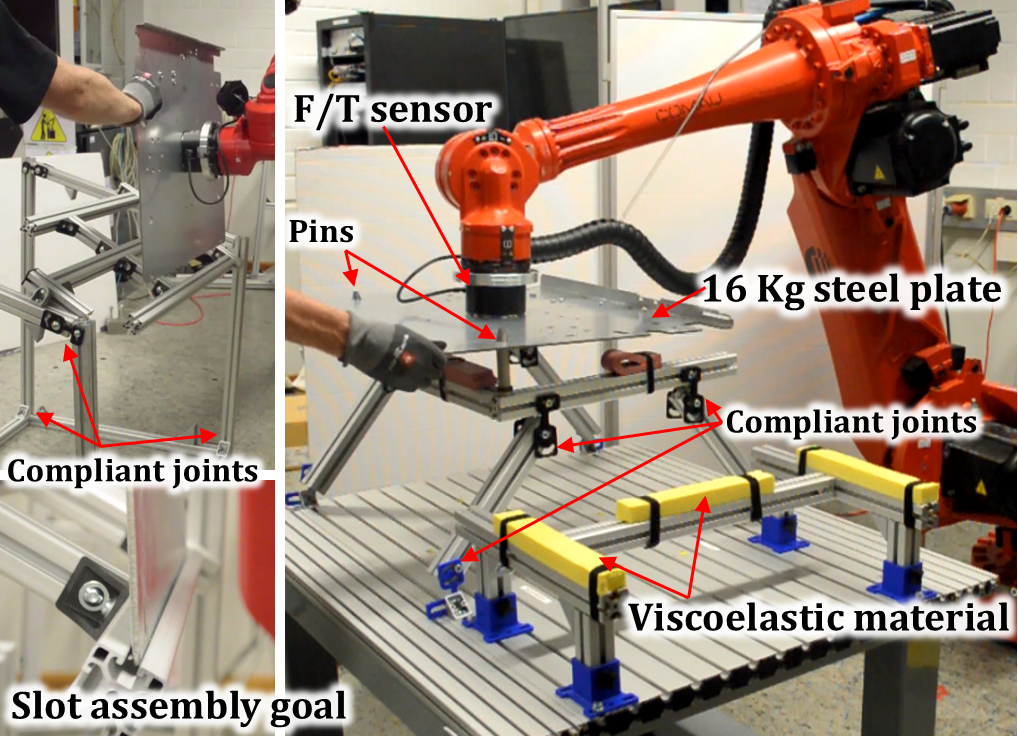} 
    \caption{Test environment for slot assembly (left), peg-in-hole (middle) and a compliant table (right) for contact tests.}
    \label{fig:contact_tests_setup}
\end{figure}

\begin{figure}
    \centering
    \includegraphics[width = .8\columnwidth]{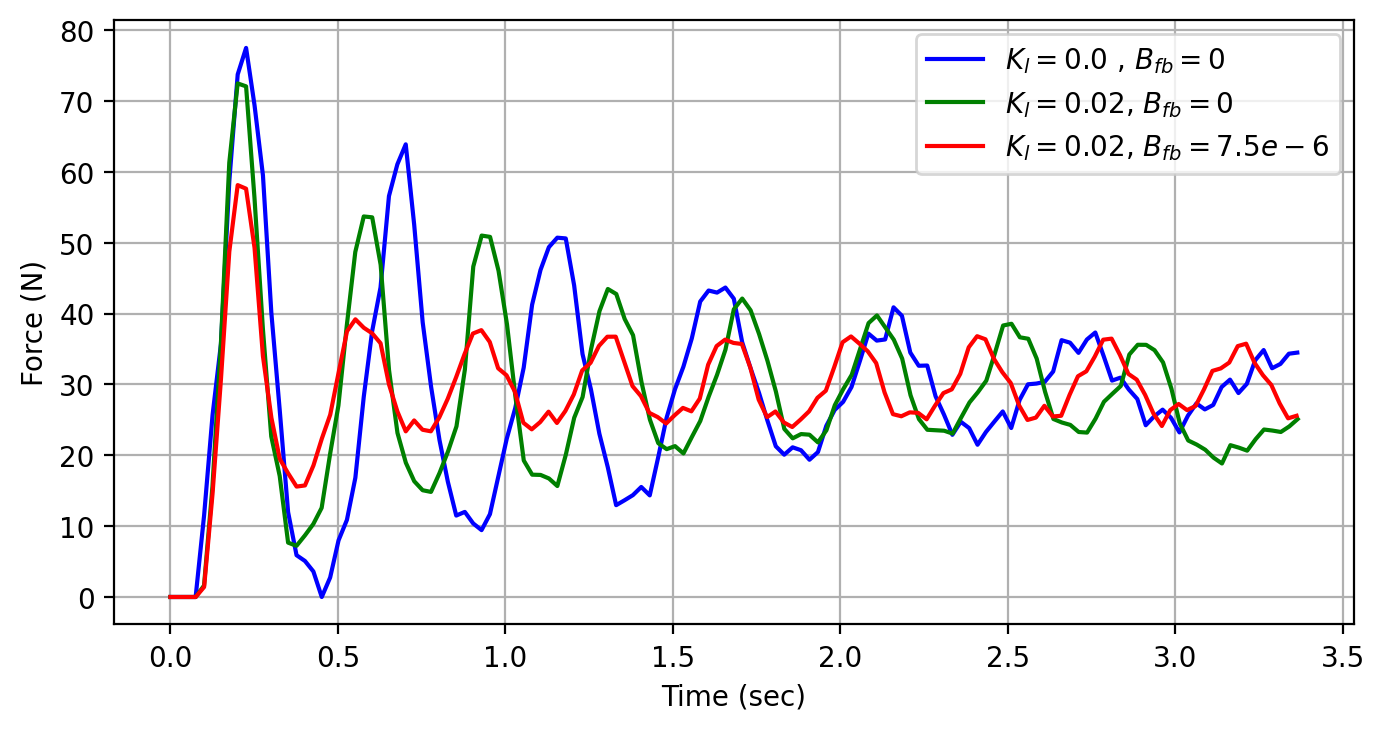}
    \caption{Impact of feedforward and feedback on contact performance with the yellow compliant table with $M_a=13$, $B_a = 1000$.}
    \label{fig:contact_tests_1}
\end{figure}

\begin{figure}
    \centering
    \includegraphics[width = .8\columnwidth]{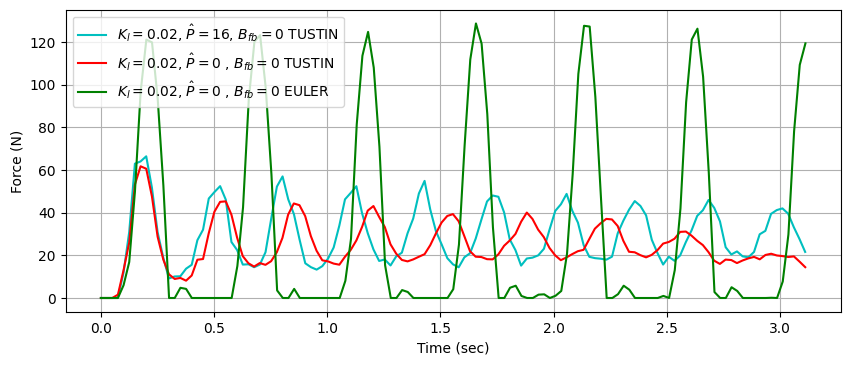} 
    \caption{The impact of acceleration compensation and discretization method on contact performance with the yellow compliant table with $M_a=10$, $B_a = 1000$.}
    \label{fig:contact_tests_3}
\end{figure}

\begin{figure}
    \centering
    \subfloat[$K_l = 0.013$, $K_e = 32$ N/mm]{\includegraphics[width = \columnwidth]{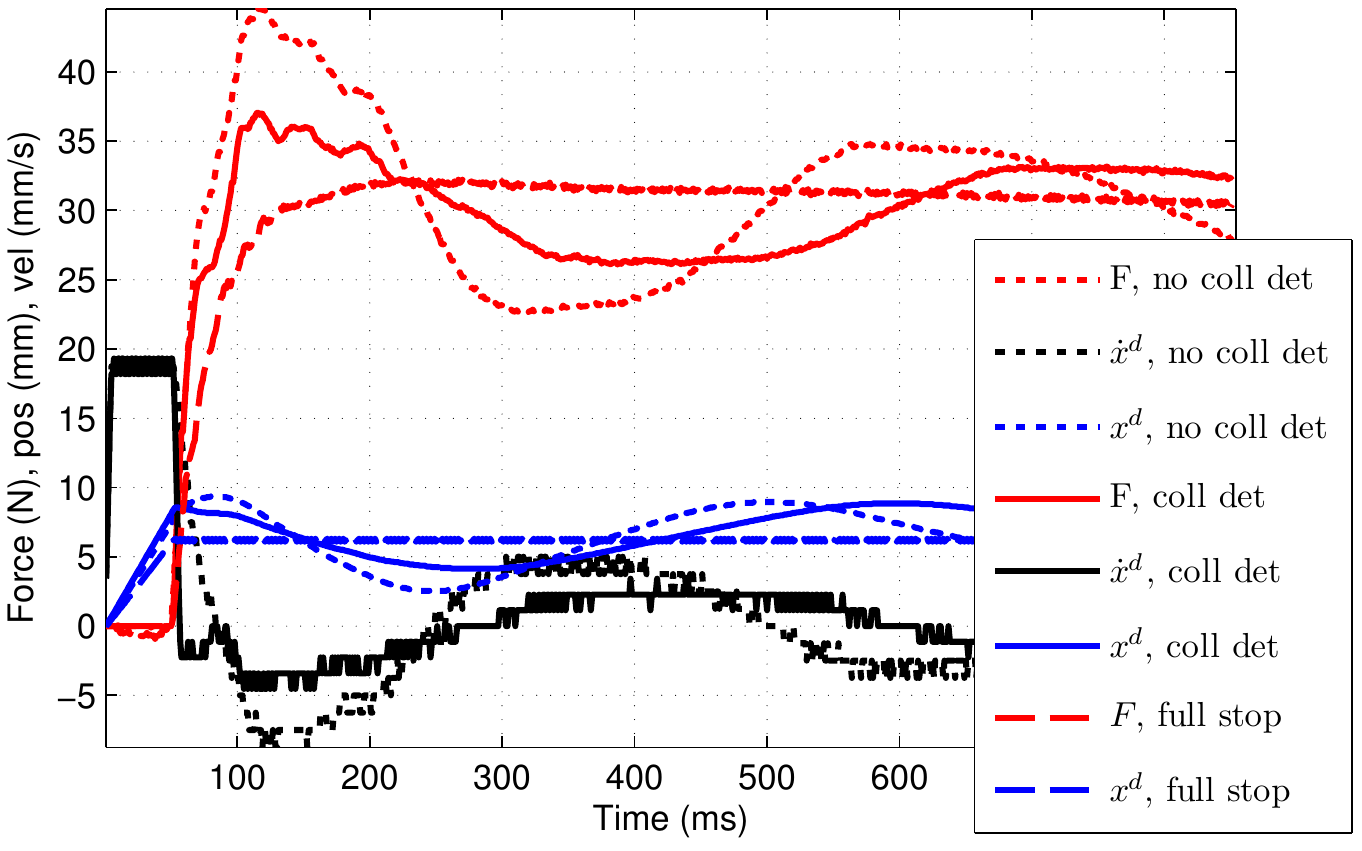}} \\
    \subfloat[$K_l = 0.020$, $K_e = 32$ N/mm]{\includegraphics[width = 0.45\columnwidth]{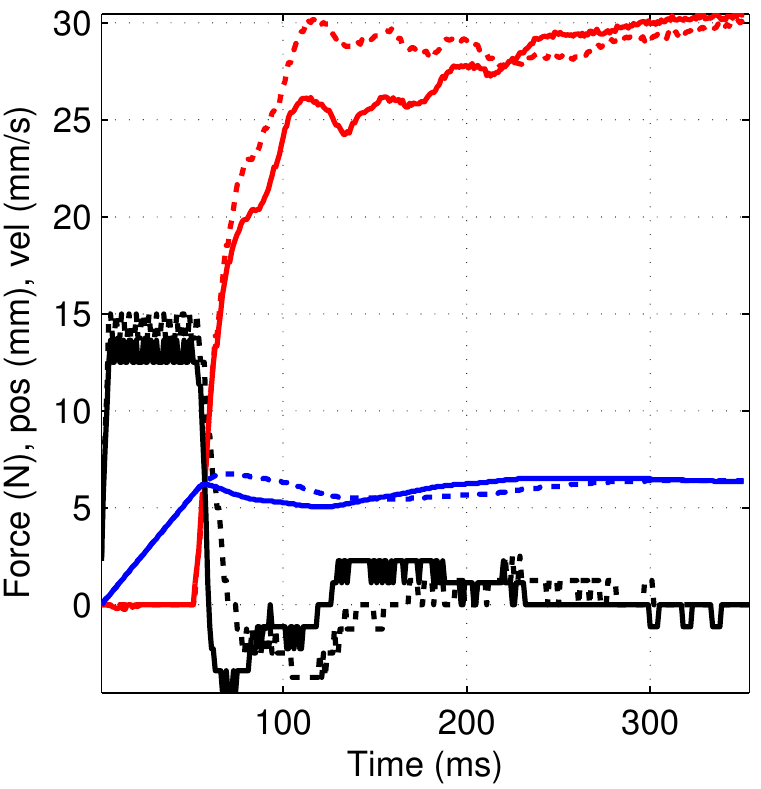}}
    \hfill
    \subfloat[$K_l = 0.013$, $K_e = 15$ N/mm]{\includegraphics[width = 0.45\columnwidth]{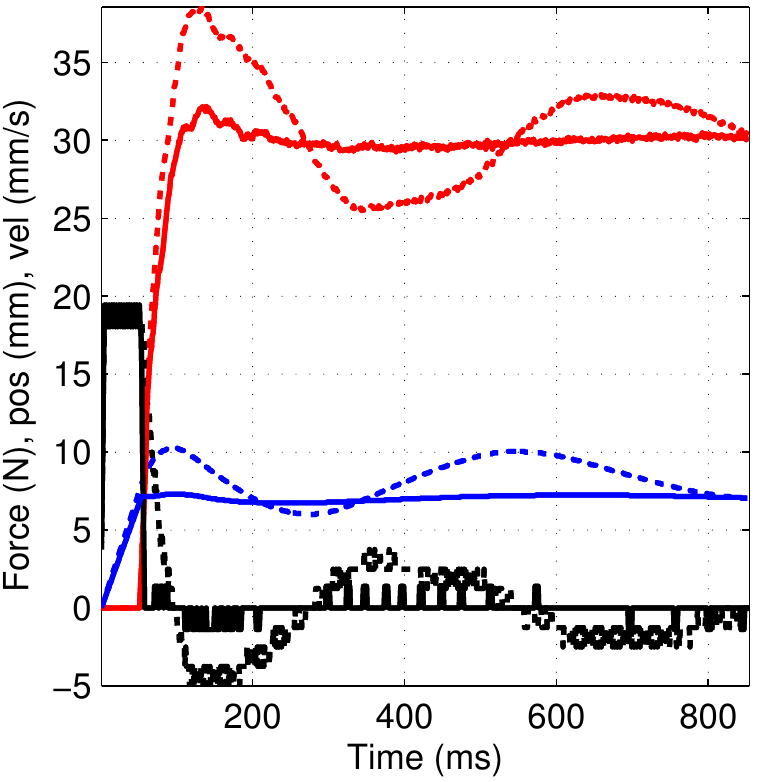}}
    \caption{\revv{Collision response method in contact, compared with standard admittance control and a full stop, at various environment stiffness $K_e$}.  Position $x^d$ is scaled 5x for visibility. Peak force is reduced by collision response, but makes less difference with higher $K_l$ and higher $K_e$.}
    \label{fig:collision_response_comparison}
\end{figure}

Compliance is critical for transient forces in contact, and is often used informally for admittance control tests on industrial robots \cite{katsura2007, ferraguti2019}. However, integrating compliance at the contact surface is not always feasible for certain tasks.  Here, we integrate compliant joints \cite{hartisch} into the table structure for the task, as can be seen in Figure \ref{fig:contact_tests_setup}. The compliant joints form a four-bar mechanism which allows the contact surface to move laterally, reducing the stiffness seen at the contact surface. The location of compliance has been shown to impact the ability to detect collision \cite{haningerc}, \rev{and a hard contact surface here can excite the payload resonance. However, integrating compliance deeper into the environment structure allows reduction in peak contact force as seen in Table I.  As seen in Fig. \ref{fig:collision_response_comparison}, the benefits of the proposed collision detection increase at lower environment stiffness.}

\begin{table}[]
    \centering
    \begin{tabular}{c|c|c|c|r|r}
        $K_l$ & $B_r$ & $\hat{M}$ (kg) & $K_e$ (N/mm) & $\max F$ (N) & $\omega$ (Hz)  \\
        \hline
         0 & 0 & 0 & $32.3$ & 77.5 & 2.1 \\
         0.02 & 0 & 0 & $32.3$ & 72.5 & 2.5 \\
         0.02 & 7.5$E^{-6}$ & 0 & $32.3$ & 58.2 & 2.6  \\
         0.02 & 7.5$E^{-6}$ & 16 & $32.3$ & 66.4 & 3.2 \\
         0.02 & 7.5$E^{-6}$ & 0 & $11.2$ & 55.6 & 3.1 \\
         0.02 & 7.5$E^{-6}$ & 0 & $5.1$ & 39.6 & 2.3 \\
         0.02 & 7.5$E^{-6}$ & 0 & $2.7$ & 34.4 & 14.7 \\
    \end{tabular}
    \caption{Control parameters, environment stiffness $K_e$, resulting maximum contact force, and frequency of dominant frequency mode $\omega$.}
    \label{tab:contact_experiments}
\end{table}

\subsection{Variable Admittance}
Adaptation of the admittance parameters has been proposed to balance free space responsiveness with good contact properties. We implement the adaptation rule proposed in \cite{ficuciello2015}, where $B_a = {\tt max}(800e^{-8|v|}, 550)$, where $v$ is robot speed in m/s. \revv{This adaptation rule is similar to many others proposed for co-manipulation \cite{abu-dakka2020} and is easily implemented.} $B_a$ is updated at 100 Hz with a limit in change of $20$ between time steps, and the admittance mass is fixed $M_a=13$. The robot is started at a position 30 cm away from contact, and an operator was asked to bring the robot into contact without knowing which control mode was active. A typical result can be seen below in Fig. \ref{fig:var_imp}. This was repeated 4 times per configuration, and one experiment dropped to bring the average contact velocity closest together. The various configurations and results can be seen in Table \ref{tab:human_contact}, evaluated according to: (1) peak contact force, (2) contact velocity, (3) total energy $\sum | F*v |$ in the direction of motion for contact, and (4) Time from first force to contact.

\begin{table}[]
    \centering
    \begin{tabular}{c|c|c|c||r|r|r|r}
        $K_l$ & $B_{fb}$ & C Det & Var $B_a$ & $\max F$ & $v_c$ & En. & Time \\
        \hline
           &  &  &  & 95.3 & 8.4 & 2.97 & 5.1 \\
         x &  &  &  & 93.9 & 7.8 & 3.05 & 5.0 \\
         x & x &  &  & 81.5 & 8.5 & 3.17 & 4.3 \\
         x & x & x &  & 76.8 & 7.8 & 3.44 & 4.2 \\
         \hline
           &  &  & x & 121 & 9.7 & 2.72 & 4.2 \\
         x & x &  & x & 81.8 & 7.8 & 3.33 & 3.5 \\
           &  & x & x & 116 & 9.3 & 2.83 & 3.4 \\
         x & x & x & x & 76.1 & 7.2 & 3.51 & 3.5 \\
    \end{tabular}
    \caption{Manual guidance to contact, where C Det is collision detection/response and Var $B_a$ is the variable admittance. $v_c$ is contact velocity (cm/s), En$ = \sum_i |F_iv_i|$ (kJ), and Time (sec) is time from start to contact.}
    \label{tab:human_contact}
\end{table}
\begin{figure}
    \centering
    \subfloat[]{\includegraphics[width=0.53\columnwidth]{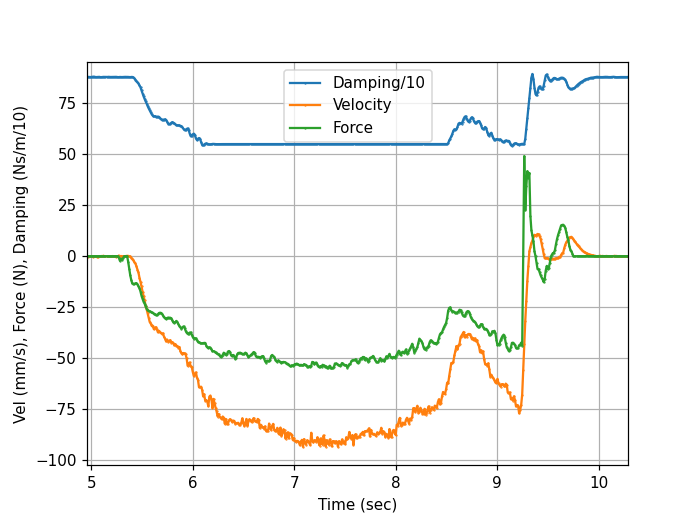}
    \includegraphics[width=0.2\columnwidth]{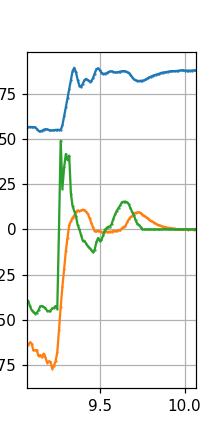}} \\
    \subfloat[]{\includegraphics[width=0.5\columnwidth]{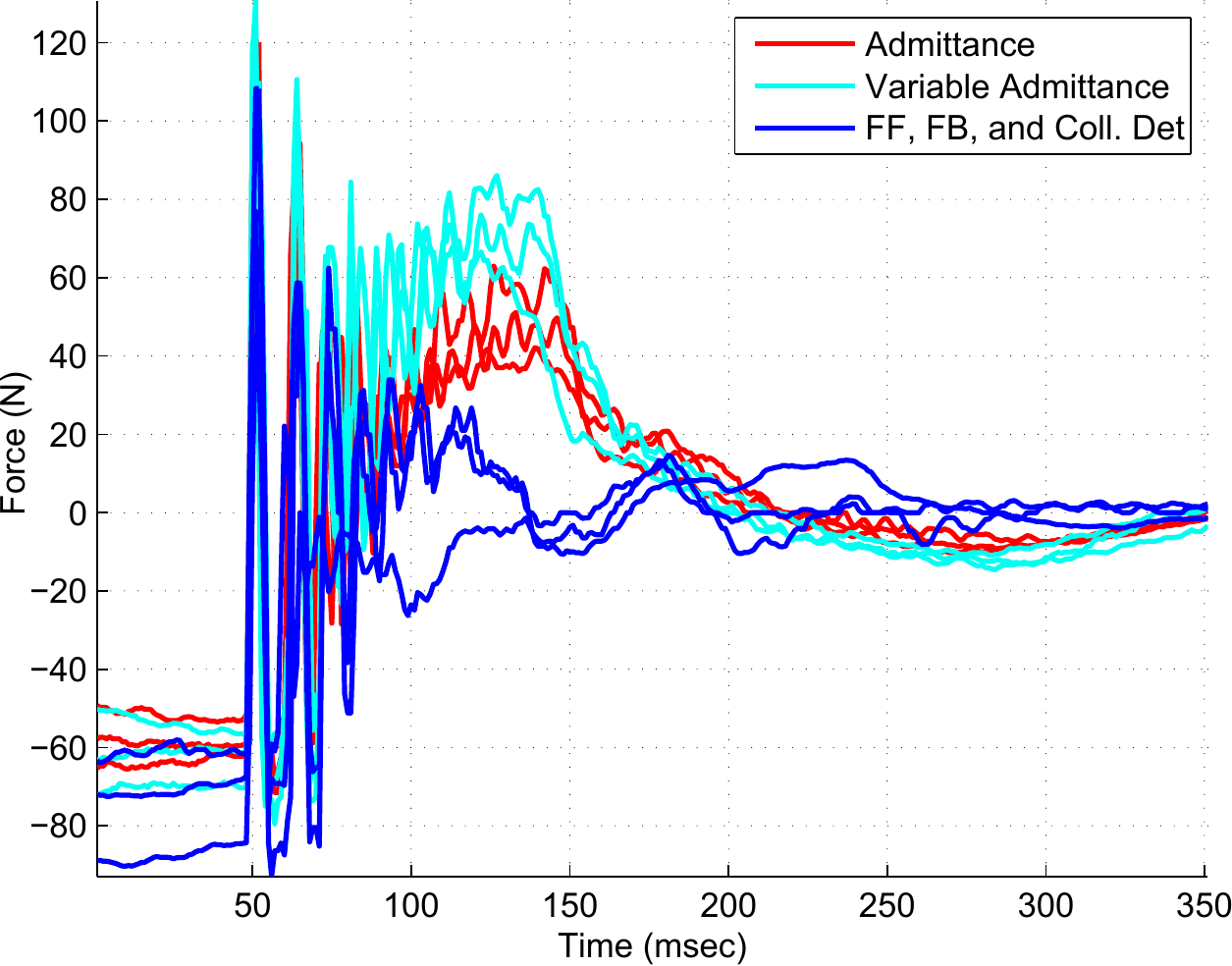} }
    \caption{Trace of experiment with variable admittance (a), and zoom of force in contact (b), comparing variable admittance, normal admittance, and the proposed controller.}
    \label{fig:var_imp}
\end{figure}

From these results, we conclude that variable impedance can improve the free-space responsiveness of the robot, reducing average completion time from $4.6$ seconds to $3.7$. However, variable admittance does not significantly improve contact performance, at least in this implementation.  As can be seen in \ref{fig:var_imp} (a, right), the damping is increased after the initial collision peak. While this result is implementation-specific, it may be difficult to address as collisions occur in 10-80 ms. Fig. \ref{fig:var_imp}(b) shows the proposed approaches improve the post-contact performance compared with standard and adaptive admittance. 




\section{Co-manipulation Demonstrations}
The design conclusions above are applied to realize high-payload co-manipulation tasks.

\subsection{16 kg Payload}
Two contact-rich assembly tasks are carried out with the setup in Figure \ref{fig:contact_tests_setup}, a double peg-in-hole assembly and a slot assembly.  In both cases, the tolerances are running (holes are oversized ~0.5mm). Compliance is integrated into the table structure, reducing total environment stiffness while still enabling the hard surface contact of the task. The linear parameters used are $K_l = 0.013$, $B_r = 7.5E^{-6}$, $\hat{P}=0$, $M_a = 13$, $B_a = 1250$ and rotational $M_a = 3$, $B_a = 150$ (rotation units are radian). For the slot assembly, rotational admittance is decreased $M_a = 12$, $B_a = 600$ to allow manipulation from one side. Stable contact transitions are achieved, as seen in the attached video \footnote{Video available at: https://youtu.be/ty-oVZs-n1Q}. 

\subsection{50 kg Payload}
The same control architecture is used with a COMAU AURA robot for a manual guidance task in free-space, seen in Figure \ref{fig:setup_man_guide}. The total payload is 50 kg, where compliance in the suction gripper system results in a payload resonance of about $9$ Hz, giving an effective $K_s\approx 1.6E^{+5}$.  Due to this resonance, the admittance gains had to be increased and $K_l$ decreased. Additionally, the payload identification routine was adapted to stop for 3 seconds before collecting measurements. Even after this modification, the payload calibration had larger magnitude of residual errors, the RMS calibration error was $23.3$ N ($1.1$ was achieved on the 16 kg payload). While the F/T sensor is rated for $5$ kN, it was factory calibrated with 10 kg test weights, resulting in more significant nonlinearities with the $50$ kg payload.

\begin{figure}
    \centering
    \includegraphics[width = .7\columnwidth]{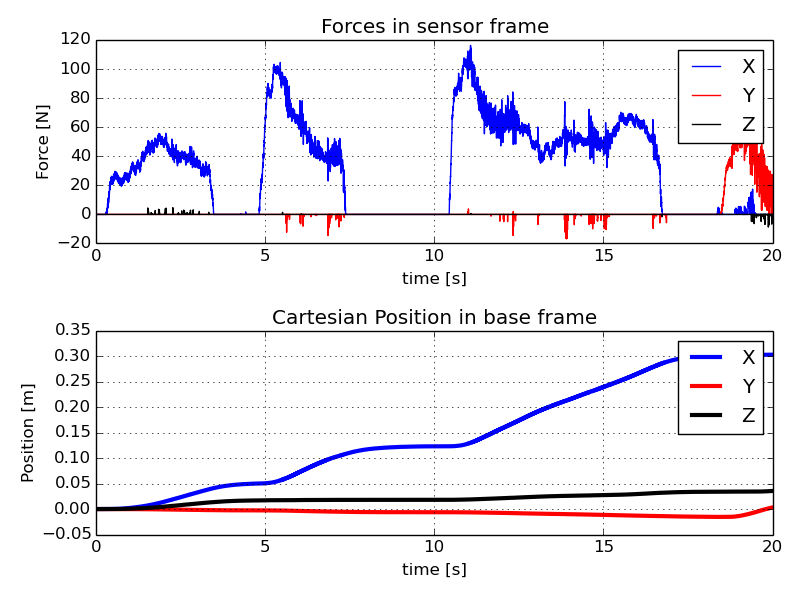} 
    \caption{Forces and Cartesian position during manual guidance with 50 kg payload.}
    \label{fig:aura_trajectory}
\end{figure}

The control parameters for this setup were tested as linear $K_l = 0.01$, $B_r = 0$, $\hat{P}=0$, $M_a = 20$, $B_a = 2000$ and rotational $M_a = 16$, $B_a = 1850$. The force clipping threshold was also increased to $20$ N and $3$ Nm.  The resulting performance can be seen in Figure \ref{fig:aura_trajectory} and the attached video. 


\section{Conclusion}
This paper took a mechatronic approach to high-payload manual guidance with environment contact. A major challenge of high-payload applications is payload resonance. \rev{The following design conclusions are drawn:

{\bf Feedforward lead $K_l$} decreases peak contact force and increasing free-space admittance. However, maximum $K_l$ is limited by payload resonance (Fig. \ref{fig:bode_plots}).

{\bf Feedback damping $B_{fb}$} decreases peak contact force (albeit less effectively than feedforward), but also reduces the admittance - this resulted in $16$\% increase in human effort on experiments here. 

{\bf Collision response} by zeroing $\dot{x}^d$ in the admittance integrator allows continuous admittance control, and reduces peak collision force by $42\%$ in Fig. 10(a). However, it is more effective with more environment compliance and requires additional integration work.

{\bf Environment compliance} reduces peak contact force (Table I) and changes the efficacy of control (Fig. 6a) and collision detection (Fig. 10b).  It can be integrated deeper in the environment structure to improve contact task feasibility (Fig. 7).

{\bf Variable damping} can reduce the time taken for a comanipulation task, but did not improve the collision characteristics here.

{\bf Payload acceleration} theoretically improves free space admittance and contact, but when the required low-pass filters are used, it increases peak contact force and reduces damping.
}


\bibliographystyle{IEEEtran}
\bibliography{lib,lib_rev,M_lib.bib}

\begin{thebibliography}{10}
\providecommand{\url}[1]{#1}
\csname url@rmstyle\endcsname
\providecommand{\newblock}{\relax}
\providecommand{\bibinfo}[2]{#2}
\providecommand\BIBentrySTDinterwordspacing{\spaceskip=0pt\relax}
\providecommand\BIBentryALTinterwordstretchfactor{4}
\providecommand\BIBentryALTinterwordspacing{\spaceskip=\fontdimen2\font plus
\BIBentryALTinterwordstretchfactor\fontdimen3\font minus
  \fontdimen4\font\relax}
\providecommand\BIBforeignlanguage[2]{{%
\expandafter\ifx\csname l@#1\endcsname\relax
\typeout{** WARNING: IEEEtran.bst: No hyphenation pattern has been}%
\typeout{** loaded for the language `#1'. Using the pattern for}%
\typeout{** the default language instead.}%
\else
\language=\csname l@#1\endcsname
\fi
#2}}

\bibitem{whitney1987}
D.~E. Whitney, ``Historical {{Perspective}} and {{State}} of the {{Art}} in
  {{Robot Force Control}},'' \emph{The International Journal of Robotics
  Research}, vol.~6, no.~1, pp. 3--14, Mar. 1987.

\bibitem{Hogan.1985_partItheory}
N.~Hogan, ``Impedance control: An approach to manipulation: Part i---theory,''
  \emph{Journal of Dynamic Systems, Measurement, and Control}, vol. 107, no.~1,
  pp. 1--7, 1985.

\bibitem{keemink2018}
A.~Q. Keemink, H.~{van der Kooij}, and A.~H. Stienen, ``Admittance control for
  physical human\textendash robot interaction,'' \emph{The International
  Journal of Robotics Research}, vol.~37, no.~11, pp. 1421--1444, Sept. 2018.

\bibitem{lecours2012}
A.~Lecours, B.~{Mayer-St-Onge}, and C.~Gosselin, ``Variable admittance control
  of a four-degree-of-freedom intelligent assist device,'' in \emph{2012
  {{IEEE}} International Conference on Robotics and Automation}.\hskip 1em plus
  0.5em minus 0.4em\relax {IEEE}, 2012, pp. 3903--3908.

\bibitem{dimeas2016}
F.~Dimeas and N.~Aspragathos, ``Online {{Stability}} in {{Human}}-{{Robot
  Cooperation}} with {{Admittance Control}},'' \emph{IEEE Transactions on
  Haptics}, vol.~9, no.~2, pp. 267--278, Apr. 2016.

\bibitem{albu-schaffer2007}
A.~{Albu-Sch{\"a}ffer}, C.~Ott, and G.~Hirzinger, ``A unified passivity-based
  control framework for position, torque and impedance control of flexible
  joint robots,'' \emph{The International Journal of Robotics Research},
  vol.~26, no.~1, pp. 23--39, 2007.

\bibitem{gosselin2013}
C.~Gosselin, \emph{et~al.}, ``A friendly beast of burden: A human-assistive
  robot for handling large payloads,'' \emph{IEEE Robotics \& Automation
  Magazine}, vol.~20, no.~4, pp. 139--147, 2013.

\bibitem{franceschi2020}
P.~Franceschi, \emph{et~al.}, ``Precise {{Robotic Manipulation}} of {{Bulky
  Components}},'' \emph{IEEE Access}, vol.~8, pp. 222\,476--222\,485, 2020.

\bibitem{roveda2018c}
L.~Roveda, \emph{et~al.}, ``Human-{{Robot Cooperative Interaction Control}} for
  the {{Installation}} of {{Heavy}} and {{Bulky Components}},'' in \emph{2018
  {{IEEE International Conference}} on {{Systems}}, {{Man}}, and
  {{Cybernetics}} ({{SMC}})}, Oct. 2018, pp. 339--344.

\bibitem{Aghili.2010}
F.~Aghili, ``Robust impedance control of manipulators carrying a heavy
  payload,'' \emph{Journal of Dynamic Systems, Measurement, and Control}, vol.
  132, no.~5, 2010.

\bibitem{audet2021}
J.-M. Audet and C.~Gosselin, ``Intuitive {{Physical Human}}-{{Robot
  Interaction}} using an {{Underactuated Redundant Manipulator}} with
  {{Complete Spatial Rotational Capabilities}},'' \emph{Journal of Mechanisms
  and Robotics}, pp. 1--15, May 2021.

\bibitem{surdilovic2020}
D.~Surdilovic, \emph{et~al.}, ``Kooperation und {{Kollaboration}} mit
  {{Schwerlastrobotern}} \textendash{} {{Sicherheit}}, {{Perspektive}} und
  {{Anwendungen}},'' in \emph{Mensch-{{Roboter}}-{{Kollaboration}}}.\hskip 1em
  plus 0.5em minus 0.4em\relax {Springer Fachmedien Wiesbaden GmbH, Edited Hans
  Buxbaum}, 2020.

\bibitem{bowyer2014a}
S.~A. Bowyer, B.~L. Davies, and F.~R. y~Baena, ``Active
  {{Constraints}}/{{Virtual Fixtures}}: A {{Survey}},'' \emph{IEEE Transactions
  on Robotics}, vol.~30, no.~1, pp. 138--157, Feb. 2014.

\bibitem{colgate1989}
E.~Colgate and N.~Hogan, ``An analysis of contact instability in terms of
  passive physical equivalents,'' in \emph{, 1989 {{IEEE International
  Conference}} on {{Robotics}} and {{Automation}}, 1989. {{Proceedings}}}, May
  1989, pp. 404--409 vol.1.

\bibitem{haddadin2017}
S.~Haddadin, A.~De~Luca, and A.~{Albu-Sch{\"a}ffer}, ``Robot collisions: A
  survey on detection, isolation, and identification,'' \emph{IEEE Transactions
  on Robotics}, vol.~33, no.~6, pp. 1292--1312, 2017.

\bibitem{haninger2019a}
K.~Haninger and D.~Surdilovic, ``Bounded {{Collision Force}} by the {{Sobolev
  Norm}}: Compliance and {{Control}} for {{Interactive Robots}},'' in
  \emph{2019 {{IEEE International Conference}} on {{Robotics}} and
  {{Automation}} ({{ICRA}})}, 2019, pp. 8259--8535.

\bibitem{sidiropoulos2021}
A.~Sidiropoulos, \emph{et~al.}, ``A variable admittance controller for
  human-robot manipulation of large inertia objects,'' in \emph{2021 30th IEEE
  International Conference on Robot \& Human Interactive Communication
  (RO-MAN)}.\hskip 1em plus 0.5em minus 0.4em\relax IEEE, 2021, pp. 509--514.

\bibitem{surdilovic2007a}
D.~Surdilovic, ``Robust control design of impedance control for industrial
  robots,'' in \emph{Intelligent {{Robots}} and {{Systems}}, 2007. {{IROS}}
  2007. {{IEEE}}/{{RSJ International Conference}} On}.\hskip 1em plus 0.5em
  minus 0.4em\relax {IEEE}, 2007, pp. 3572--3579.

\bibitem{cao2020}
H.~Cao, \emph{et~al.}, ``Smooth adaptive hybrid impedance control for robotic
  contact force tracking in dynamic environments,'' \emph{Industrial Robot: the
  international journal of robotics research and application}, vol.~47, no.~2,
  pp. 231--242, Jan. 2020.

\bibitem{S.Farsoni.2017}
{S. Farsoni}, \emph{et~al.}, ``Compensation of load dynamics for admittance
  controlled interactive industrial robots using a quaternion-based kalman
  filter,'' \emph{IEEE Robotics and Automation Letters}, vol.~2, no.~2, pp.
  672--679, 2017.

\bibitem{katsura2007}
S.~Katsura, Y.~Matsumoto, and K.~Ohnishi, ``Modeling of force sensing and
  validation of disturbance observer for force control,'' \emph{IEEE
  Transactions on industrial electronics}, vol.~54, no.~1, pp. 530--538, 2007.

\bibitem{pham2020}
H.~Pham and Q.-C. Pham, ``Convex {{Controller Synthesis}} for {{Robot
  Contact}},'' \emph{IEEE Robotics and Automation Letters}, vol.~5, no.~2, pp.
  3330--3337, Apr. 2020.

\bibitem{abu-dakka2020}
F.~J. {Abu-Dakka} and M.~Saveriano, ``Variable impedance control and learning
  -- {{A}} review,'' \emph{arXiv:2010.06246 [cs]}, Oct. 2020.

\bibitem{ficuciello2015}
F.~Ficuciello, L.~Villani, and B.~Siciliano, ``Variable impedance control of
  redundant manipulators for intuitive human\textendash robot physical
  interaction,'' \emph{IEEE Transactions on Robotics}, vol.~31, no.~4, pp.
  850--863, 2015.

\bibitem{ferraguti2019}
F.~Ferraguti, \emph{et~al.}, ``A variable admittance control strategy for
  stable physical human\textendash robot interaction,'' \emph{The International
  Journal of Robotics Research}, vol.~38, no.~6, pp. 747--765, May 2019.

\bibitem{tsumugiwa2002}
T.~Tsumugiwa, R.~Yokogawa, and K.~Hara, ``Variable impedance control based on
  estimation of human arm stiffness for human-robot cooperative calligraphic
  task,'' in \emph{Robotics and {{Automation}}, 2002. {{Proceedings}}.
  {{ICRA}}'02. {{IEEE International Conference}} On}, vol.~1.\hskip 1em plus
  0.5em minus 0.4em\relax {IEEE}, 2002, pp. 644--650.

\bibitem{colgate1997}
J.~E. Colgate and G.~G. Schenkel, ``Passivity of a class of sampled-data
  systems: Application to haptic interfaces,'' \emph{Journal of robotic
  systems}, vol.~14, no.~1, pp. 37--47, 1997.

\bibitem{hogan1988}
N.~Hogan, ``On the stability of manipulators performing contact tasks,''
  \emph{Robotics and Automation, IEEE Journal of}, vol.~4, no.~6, pp. 677--686,
  1988.

\bibitem{haninger2018b}
K.~Haninger and D.~Surdilovic, ``Identification of {{Human Dynamics}} in
  {{User}}-{{Led Physical Human Robot Environment Interaction}},'' in
  \emph{2018 27th {{International Symposium}} on {{Robot}} and {{Human
  Interactive Communication}} ({{RO}}-{{MAN}})}, 2018, pp. 509--514.

\bibitem{eppinger1992}
S.~D. Eppinger and W.~P. Seering, ``Three dynamic problems in robot force
  control,'' \emph{IEEE Transactions on Robotics and Automation}, vol.~8,
  no.~6, pp. 751--758, 1992.

\bibitem{buerger2006}
S.~P. Buerger and N.~Hogan, ``Relaxing {{Passivity}} for {{Human}}-{{Robot
  Interaction}},'' in \emph{2006 {{IEEE}}/{{RSJ International Conference}} on
  {{Intelligent Robots}} and {{Systems}}}, Oct. 2006, pp. 4570--4575.

\bibitem{haningerc}
K.~Haninger, \emph{et~al.}, ``Contact {{Information Flow}} and {{Design}} of
  {{Compliance}},'' in \emph{On {{Review}}, {{ICRA}} 2022}.

\bibitem{okunev2012}
V.~Okunev, T.~Nierhoff, and S.~Hirche, ``Human-preference-based control design:
  Adaptive robot admittance control for physical human-robot interaction,'' in
  \emph{{{RO}}-{{MAN}}, 2012 {{IEEE}}}.\hskip 1em plus 0.5em minus 0.4em\relax
  {IEEE}, 2012, pp. 443--448.

\bibitem{hartisch}
R.~Hartisch and K.~Haninger, ``Co-design of {{Environmental Compliance}} for
  {{High}}-speed {{Contact Tasks}},'' in \emph{On {{Review}}, 2021}.

\end{thebibliography}

\end{document}